\documentclass{article}

\usepackage{mlps_conf}
\usepackage{times}
\usepackage{epsfig}
\usepackage{graphicx}
\usepackage{amsmath}
\usepackage{amssymb}
\usepackage{pgfplots}
\pgfplotsset{compat=1.18}
\usepackage{xcolor}
\usepackage{booktabs}
\usepackage{enumitem}
\usepackage{pifont}

\usepackage[utf8]{inputenc} % allow utf-8 input
\usepackage[T1]{fontenc}    % use 8-bit T1 fonts
\usepackage{url}            % simple URL typesetting
\usepackage{amsfonts}       % blackboard math symbols
\usepackage{nicefrac}       % compact symbols for 1/2, etc.
\usepackage{microtype}      % microtypography
\usepackage{fontawesome5}
\usepackage{tabularx}
\usepackage{subcaption}

\definecolor{myGray}{rgb}{0.5, 0.5, 0.5}
\definecolor{myRed}{rgb}{0.808,0.067,0.149}
\definecolor{myGreen}{rgb}{0.067,0.708,0.149}

\usepackage{multirow}

\usepackage{longtable}
\usepackage[skins]{tcolorbox} 
\tcbuselibrary{breakable}
\usepackage{float}
\newfloat{Prompt}{htbp}{loa}
\usepackage{placeins}
\usepackage{times}
\usepackage{epsfig}
\usepackage{graphicx}
\usepackage{amsmath}
\usepackage{amssymb}
\definecolor{darkgreen}{rgb}{0.0,0.5,0.0}
\usepackage{colortbl}
\usepackage{enumitem}

\usepackage{tikz}
\usetikzlibrary{positioning,arrows.meta,fit,calc,shapes.geometric}
\usepackage{graphicx}

\usepackage[rightcaption]{sidecap}
\usepackage{graphicx}

\copyrightnotice{979-8-3503-2411-2/25/\$31.00 {\copyright}2025 IEEE}

\usepackage[pagebackref=true,breaklinks=true,letterpaper=true,colorlinks,bookmarks=false]{hyperref}

\name{
\parbox{\textwidth}{\centering
David Robinson$^{1}$ \quad
Animesh Gupta$^{1}$ \quad
Elizabeth Clark$^{3}$ \\
Olga Melnik$^{3}$ \quad
Qiushi Fu$^{2}$ \quad
Mubarak Shah$^{1}$
}
}
\address{
$^{1}$Center for Research in Computer Vision, University of Central Florida \\
$^{2}$Mechanical and Aerospace Engineering, University of Central Florida \\
$^{3}$Doctor of Physical Therapy Department, AdventHealth University
}

\title{Enhancing Box and Block Test with Computer Vision for Post-Stroke Upper Extremity Motor Evaluation}

\begin{document}

%%%%%%%%% TITLE

\maketitle

\begin{abstract}

Standard clinical assessments of upper-extremity motor function after stroke either rely on ordinal scoring, which lacks sensitivity, or time-based task metrics, which do not capture movement quality. In this work, we present a computer vision-based framework for analysis of upper-extremity movement during the Box and Block Test (BBT) through world-aligned joint angles of fingers, arm, and trunk without depth sensors or calibration objects. We apply this framework to a dataset of 136 BBT recordings collected from 48 healthy individuals and 7 individuals post stroke. Using unsupervised dimensionality reduction of joint-angle features, we analyze movement patterns without relying on expert clinical labels. The resulting embeddings show separation between healthy movement patterns and stroke-related movement deviations. Importantly, some patients with the same BBT scores can be separated with different postural patterns. These results show that world-aligned joint angles can capture meaningful information of upper-extremity functions beyond standard time-based BBT scores, with no effort from the clinician other than monocular video recordings of the patient using a phone or camera. This work highlights the potential of a camera-based, calibration-free framework to measure movement quality in clinical assessments without changing the widely adopted clinical routine.

\end{abstract}

\begin{figure*}[t] % Use [H] to force exact placement; [htbp] is more flexible
    \centering
    \includegraphics[width=\textwidth]{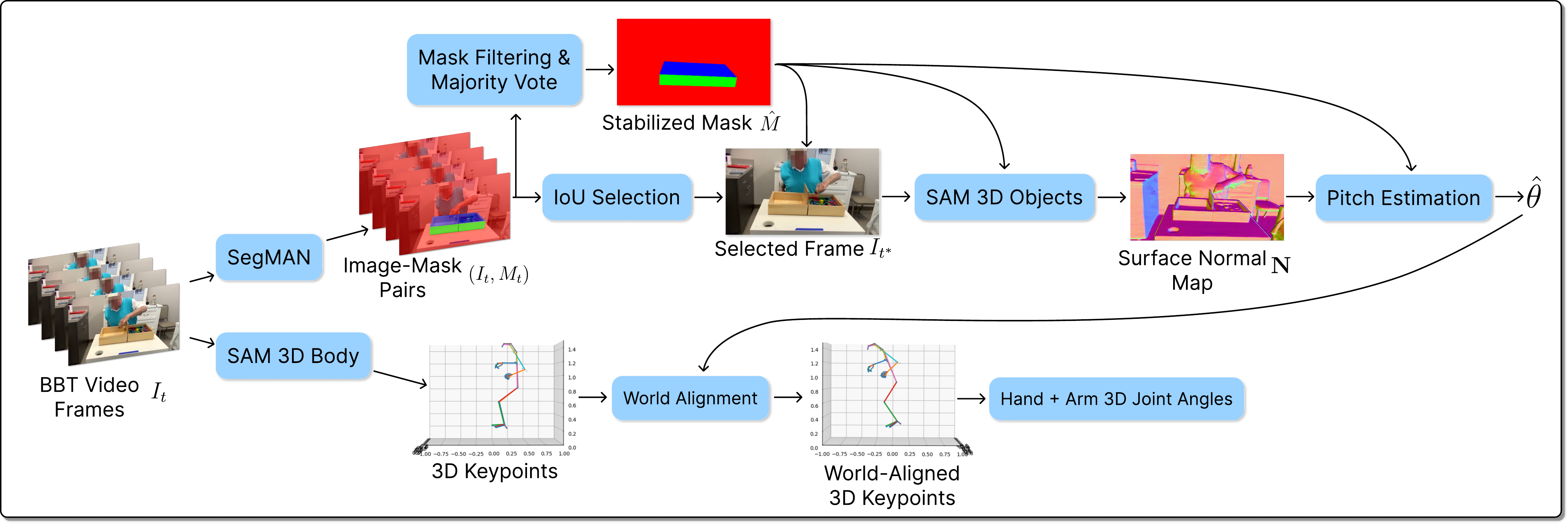} % Adjust width and filename
    \caption{Overview of our world-aligned 3D joint angle estimation pipeline. Video frames from the Box and Block Test are segmented to recover the box orientation, which is used to estimate camera pitch through surface normals. Monocular 3D keypoints are then extracted and rotated to align with gravity, enabling measurement of finger and arm joint angles independent of camera orientation.}
    \label{fig:main}
  \vspace{-1em}
\end{figure*}

\section{Introduction}
\label{sec:section1}
Stroke is the leading cause of serious chronic physical disability worldwide \cite{Cieza2020Lancet}, with 50\%-80\% of acute stroke survivors exhibiting some Upper Extremity (UE) dysfunction \cite{Lawrence2001Stroke}, and 30\%-66\% remains with UE impairments \cite{Kwakkel2003Stroke}. Despite advances in treatment options, critical limitations remain in the clinical assessment of UE function for rehabilitation. Most existing assessment instruments can be categorized as one of the following methods: (1) using an ordinal scale for qualitative evaluation across a wide range of tasks (e.g., FMA-UE \cite{Fugl1975SJRM}, ARAT \cite{Lyle1981IJRR}), (2) using a time-based quantitative evaluation of a specific task (e.g., BBT \cite{Mathiowetz1985AJOT}, NHPT \cite{Kellor1971AJOT}), or a combination of the two (WMFT \cite{Wolf2001Stroke}). The former method can assess movement quality comprehensively over multiple joints/tasks, but it has limited sensitivity due to its discrete subjective scoring levels. More importantly, these multi-item instruments can take a relatively long time for well-trained therapists to administer. In contrast, the latter method is fast to use, but it only generates task completion time without movement quality information. Therefore, existing assessment instruments cannot be effectively used in routine rehabilitation sessions to assist therapists in evidence-based optimization of interventions.

The objective of the present study is to develop and validate an objective UE movement assessment framework by integrating state-of-the-art computer vision (CV) algorithms with standard time-based assessment tools. In recent years, CV has been frequently used to automate the scoring of existing instruments \cite{Simbana2019IEEE}, especially with the help of RGB-D cameras (e.g., Microsoft Kinect). For instance, \cite{Hsiao2013IEEE, Ona2017SIP} count the number of blocks automatically in BBT, \cite{Kim2016PLOSONE, Lee2018IEEE, Olesh2014PLOSONE, Simonsen2017JNER} predicted FMA-UE, ARAT, and JTHF scores across various items. However, such automation does not reduce the time of administering these tests nor provide any additional information beyond the original scores. In contrast, CV-based skeletal pose estimation has been used in rehabilitation interventions to quantify compensatory movements via both RGB-D or standard cameras \cite{Averell2022, Cai2024RAS, Coias2022JNER}, but these interventions are not standardized assessment tools to be used regularly by physicians or therapists.

The present study builds on our recent research that used CV to extract 2D skeletal poses from a regular low-cost camera to detect action phases in the clinically well-established BBT (Box and Block Test) assessment \cite{Robinson2025MLSP}. We implemented state-of-the-art 3D skeletal pose estimation algorithms to estimate UE joint angles (trunk, arm, and fingers) from single-camera (monocular) video recordings of BBT. This approach enabled us to provide rich movement quality information in addition to standard BBT scores with minimal interference to the existing clinical assessment procedure. Our preliminary data with healthy individuals and 7 stroke survivors demonstrated the potential benefits of this approach, allowing us to differentiate patients based on their unique postural characteristics.

In summary, we make the following contributions in this work:
\begin{itemize}
    \item A camera-invariant, minimally intrusive framework for upper-extremity movement analysis that estimates world-aligned 3D joint angles from monocular RGB video of Box and Block Test by recovering camera pitch through semantic segmentation and surface normal estimation of the test apparatus.
    \item A qualitative evaluation of state-of-the-art monocular 3D body and hand pose estimation methods for the Box and Block Test, motivating the selection of a pose estimation method for robust 3D joint-angle estimation under occlusion.
\end{itemize}

% \begin{figure*}[t] % Use [H] to force exact placement; [htbp] is more flexible
%     \centering
%     \includegraphics[width=\textwidth]{figures/fig1.png} % Adjust width and filename
%     \caption{}
%     \label{fig:example_figure}
% \end{figure*}

% \section{Current 3D Pose Methods}
% \input{section2}

\begin{figure*}[t]
  \centering
  \includegraphics[width=\linewidth]{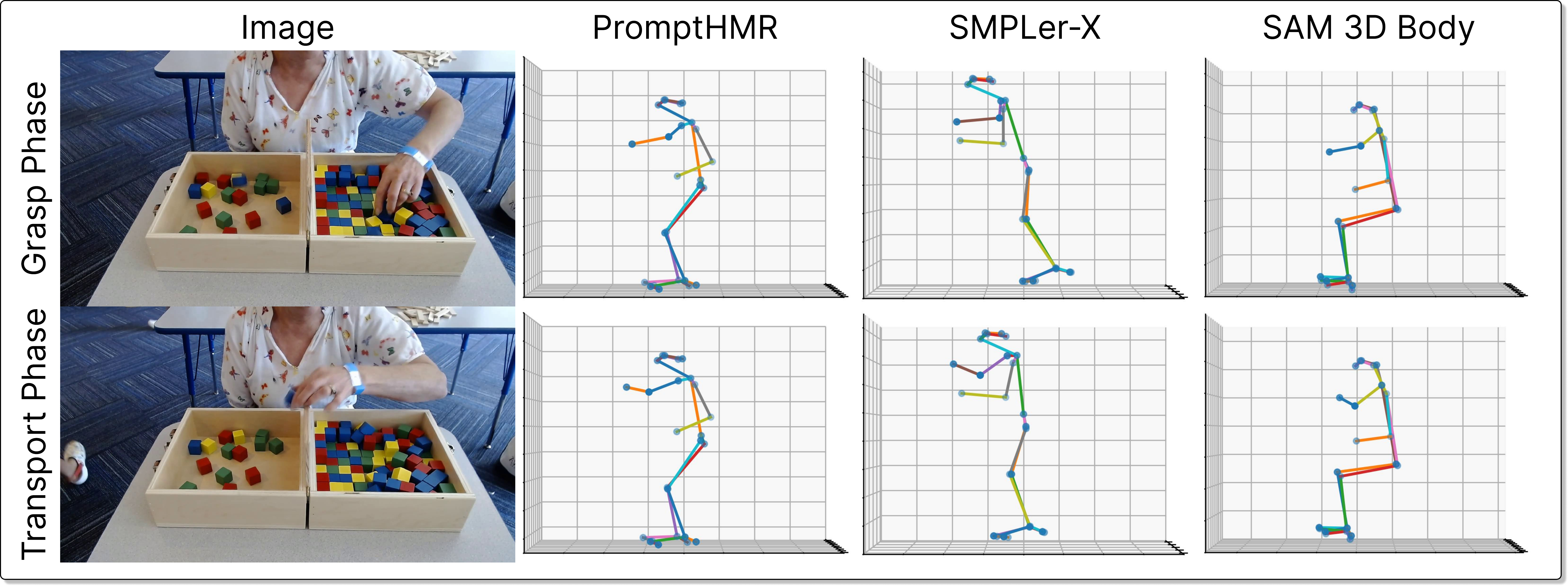}
  \caption{Comparison of 3D body pose estimation methods during the grasp and transport phases of the Box and Block Test. Side-view visualizations of 3D keypoints are shown for PromptHMR, SMPLer-X, and SAM 3D Body, illustrating differences in depth consistency and joint articulation relative to the observed arm pose in the image.}
  \label{fig:body_pose}
  \vspace{-1em}
\end{figure*}

\section{Method}
\subsection{Recording Box and Block Tests}

We recorded videos of 136 Box and Block Tests from 48 healthy individuals (age 5-73 years, 20 Males) and 7 individuals post stroke (see Table \ref{tab:patient_scores}). For healthy individuals, each generated one BBT recording per hand. For the group diagnosed with stroke, each participant generated three BBT recordings per hand over three visits (except one patient) that were separated by one week. The experimental protocol was approved by the Institutional Review Board at AdventHealth University and the University of Central Florida. In each recording, the subject repeatedly transfers blocks from one box to another as fast as they can within one minute. All videos were recorded at 1080p resolution and standardized to 22 FPS. The faces of the participants were either blurred or excluded from the recordings for privacy protection.

\subsection{Zero-Shot Monocular Extrinsic Camera Calibration}

We perform monocular extrinsic camera calibration by segmenting the BBT box across video frames using a fine-tuned SegMAN \cite{SegMAN} model, filtering and aggregating frame-level masks to obtain a temporally stabilized box mask and a representative frame. This stabilized mask and selected frame are used to reconstruct a 3D point map through SAM 3D Objects \cite{sam3dteam2025sam3d3dfyimages}, from which surface normals are estimated and aggregated over the box's front face to recover the camera pitch relative to gravity.

\textbf{SegMAN fine-tuning:} For each of the 55 subjects, we sampled 1 frame without occlusion and 14 frames with hand-occlusion, resulting in $55\times 15 = 825$ images. We annotated a single segmentation mask between the front of the box, rest of the box, and background for each video, and that mask serves as the ground-truth for each of the 15 frames per video. Applying an 80/20 train-test split on the original videos to prevent data leakage, we fine-tuned SegMAN \cite{SegMAN} for semantic segmentation on $44\times 15=660$ images and tested on the remaining $11\times 15=165$ images, resulting in a 0.9492 mIoU on the test set. This level of accuracy was sufficient to yield stable surface normal estimates in all videos tested. Given an RGB input frame $I$, our fine-tuned SegMAN \cite{SegMAN} produces a mask $M\in{\{0,1,2\}}^{H\times W}$ between the background, front of the BBT box, and the rest of the box (including areas occluded by the hand), respectively. We then derive a binary box mask by thresholding the multi-class mask.
\[
M^\text{box}_{u,v} =\begin{cases}
1 & M_{u,v} > 0 \\
0 & M_{u,v} \leq 0
\end{cases}
\]

We apply SegMAN \cite{SegMAN} to each frame of the video to form the set of image-mask pairs.
\[
\mathcal{D}=\{(I_1, M_1),(I_2, M_2),\ldots,(I_T, M_T)\}
\]
 
\textbf{Mask filtering:} To reduce noise and enforce temporal consistency, we discard masks that have multiple disconnected box clusters or mask outliers relative to the median box area. Let $CC(\cdot)$ return the set of connected components for the foreground in a binary mask, and let $|CC(\cdot)|$ be the number of components.
\[
\tilde{M}^\text{box}_{u,v}=\operatorname{median}\{M_{t,u,v}^\text{box}\;\mid\; (I_t,M_t)\in \mathcal{D}\}
\]
\[
\mathcal{P}(M): \operatorname{IoU}(M^\text{box}, \tilde{M}^\text{box})\geq \tau,\; |CC(M^\text{box})| = 1
\]
where $\tilde{M}^\text{box}$ denotes the pixel-wise median of the binary masks from $\mathcal{D}$, $\tau$ is a hyperparameter, and $CC(\cdot)$ denotes the number of connected box components under 8-connectivity.
\[
\mathcal{D}^\text{filt}=\{(I_t,M_t)\in\mathcal{D} \;\mid\;\mathcal{P}(M_t)\}
\]

\textbf{Majority vote:} The majority vote for each pixel from the masks in $\mathcal{D}^\text{filt}$ is taken to form the final mask $\hat{M}$, representing a temporally stabilized estimate of the box mask. We select the frame whose mask has the highest IoU with $\hat{M}^\text{box}$.
\[
t^*=\arg\limits\max_{t\in\{1,\ldots,T\}} \operatorname{IoU}(M_t^\text{box},\hat{M}^\text{box})
\]

\textbf{Surface normal estimation:} The corresponding image $I_{t^*}$ and mask $\hat{M}^\text{box}$ are passed to SAM 3D Objects \cite{sam3dteam2025sam3d3dfyimages} together to generate a 3D point cloud of the object from the camera perspective as a point map $P\in\mathbb{R}^{H\times W\times 3}$. Given the point map $P$ returned by SAM 3D Objects \cite{sam3dteam2025sam3d3dfyimages}, we estimate a per-pixel surface normal field $\mathbf{N}_{u,v}\in\mathbb{R}^{H\times W\times 3}$ using local finite differences in the image grid. For each valid pixel $(u,v)$ where finite differences are defined, we compute tangent vectors
\[
\Delta_u P_{u,v}=P_{u+1,v}-P_{u,v},\quad \Delta_v P_{u,v}=P_{u,v+1} - P_{u,v},
\]
and define the unit normal as the normalized cross product
\[
\mathbf{N}_{u,v}=\frac{\Delta_v P_{u,v}\times \Delta_u P_{u,v}}{{\|\Delta_v P_{u,v}\times \Delta_u P_{u,v}\|}_2 + \epsilon}
\]
where $\epsilon={10}^{-9}$ is a small constant for numerical stability.

\textbf{Camera pitch estimation:} We convert each normal $\mathbf{}_{u,v}={[n_x,n_y,n_z]}^\top$ into a pitch angle relative to the camera.
\[
\theta_{u,v}=\arctan2(n_z,-n_y)
\]
Let $M^\text{front}\in {\{0,1\}}^{H\times W}$ represent the binary mask for the front face of the box. The final pitch estimate is computed by taking the median of the pitch angles of the masked pixels.
\[
\hat{\theta}=\operatorname{median}\{\theta_{u,v} \mid M^\text{front}_{u,v}=1\}
\]

\begin{table*}[t]
\centering
\footnotesize
\begin{tabular}{lccccccccc}
\toprule
& \multicolumn{3}{c}{3DPW \cite{3dpw}} &
  \multicolumn{3}{c}{EMDB \cite{kaufmann2023emdb}} &
  \multicolumn{3}{c}{RICH \cite{rich}} \\
\cmidrule(lr){2-4}
\cmidrule(lr){5-7}
\cmidrule(lr){8-10}
Method
& PA-MPJPE & MPJPE & PVE
& PA-MPJPE & MPJPE & PVE
& PA-MPJPE & MPJPE & PVE \\
\midrule
PromptHMR \cite{wang2025prompthmr} & 36.1 & 58.7 & 69.4 & 41.0 & 71.7 & 84.5 & 37.3 & 56.6 & 65.5 \\
SMPLer-X \cite{cai2023smplerx} & 46.6 & 76.7 & 91.8 & 64.5 & 92.7 & 112.0 & 37.4 & 62.5 & 69.5 \\
SAM 3D Body \cite{yang2025sam3dbody} & \textbf{33.8} & \textbf{54.8} & \textbf{63.6} & \textbf{38.2} & \textbf{61.7} & \textbf{72.5} & \textbf{30.9} & \textbf{53.7} & \textbf{60.3} \\
\bottomrule
\end{tabular}
\caption{Comparison of body pose estimation methods on three common benchmarks where the best results are highlighted in bold. Results for SMPLer-X \cite{cai2023smplerx} are taken from the SAM 3D Body benchmark using publicly released checkpoints.}
\label{tab:body_pose}
\vspace{-1.5em}
\end{table*}

\subsection{Monocular 3D Pose Estimation}

Pose estimation is performed independently per frame, without temporal filtering. The whole RGB frame is passed through SAM 3D Body \cite{yang2025sam3dbody}, which outputs a Momentum Human Rig (MHR) parametric model, including hand and body keypoints, in the camera coordinate frame. The keypoints are filtered for upper-extremity movement, including finger, wrist, elbow, shoulder, and torso keypoints. Using the previous camera pitch estimation method, we rotate the keypoints to align with gravity direction and account for camera pitch. Gravity alignment is important for separating true posture from apparent trunk orientation caused by camera tilt. Finger, wrist, elbow, shoulder, and trunk angles are computed as the 3D angle between adjacent limb vectors as triplets of joints. 

\section{Results}
\subsection{Comparison of 3D Pose Estimation Methods}

\begin{figure*}[t]
  \centering
  \begin{subfigure}[b]{0.3\textwidth}
    % include the barplot TikZ figure
    \includegraphics[width=\textwidth]{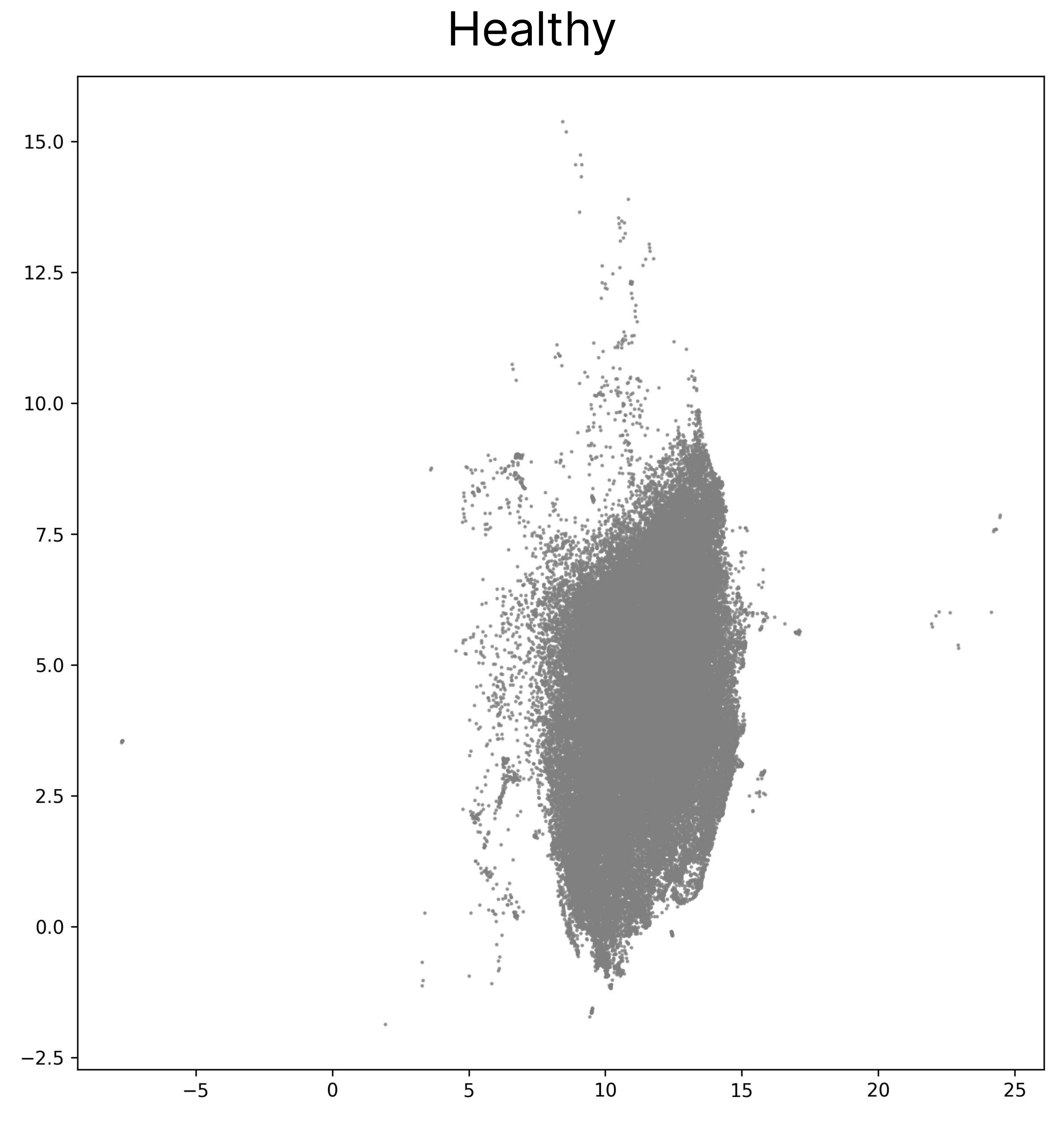}
    \caption{Joint angle embeddings of healthy subjects only.\vspace{1.5em}}
    \label{fig:healthy_umap}
  \end{subfigure}
  \hfill
  \begin{subfigure}[b]{0.3\textwidth}
    % include the barplot TikZ figure
    \includegraphics[width=\textwidth]{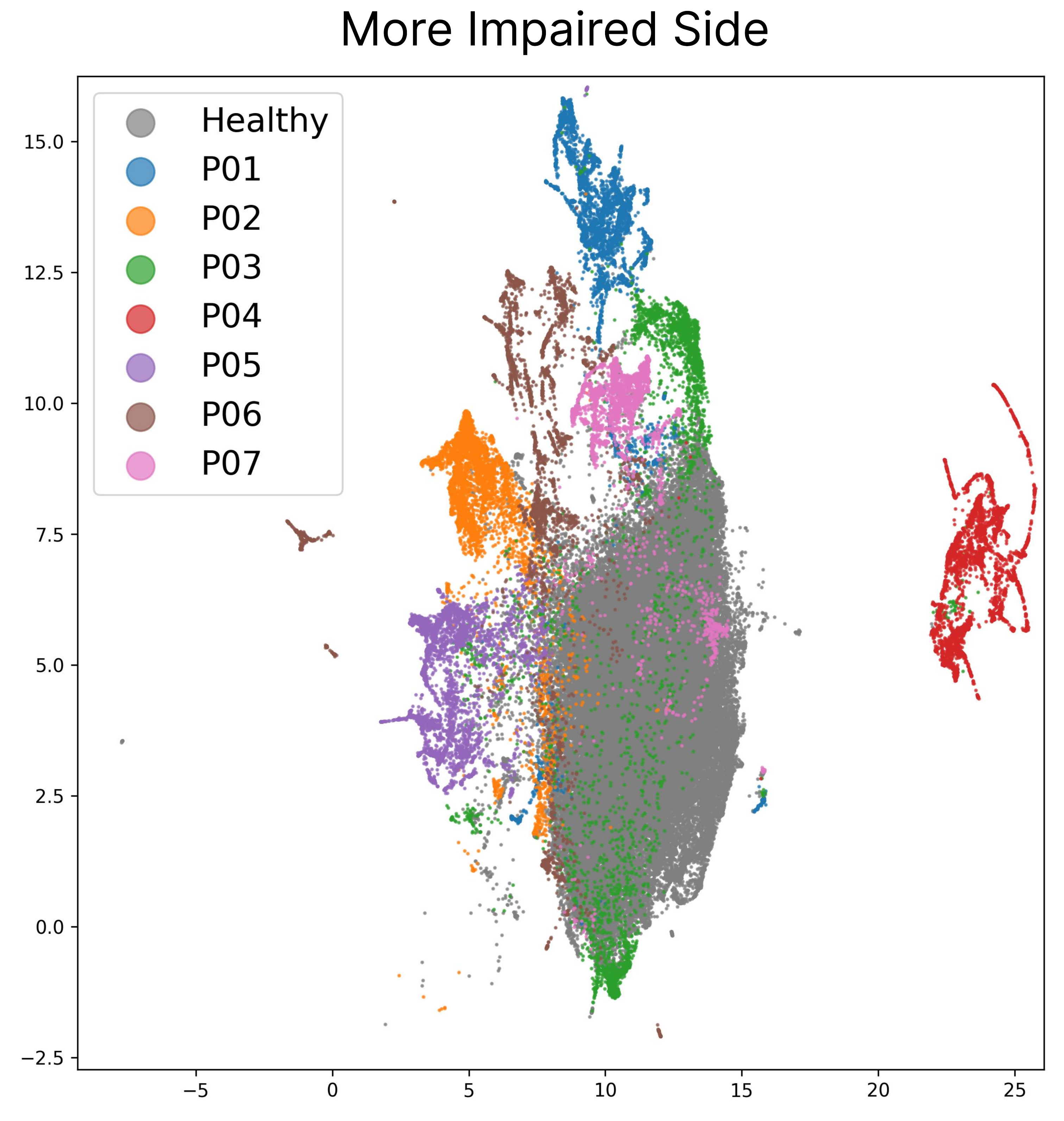}
    \caption{Healthy subject embeddings overlaid with joint-angle embeddings from each patient's more impaired side.}
    \label{fig:mi_umap}
  \end{subfigure}
  \hfill
  \begin{subfigure}[b]{0.3\textwidth}
    % include the barplot TikZ figure
    \includegraphics[width=\textwidth]{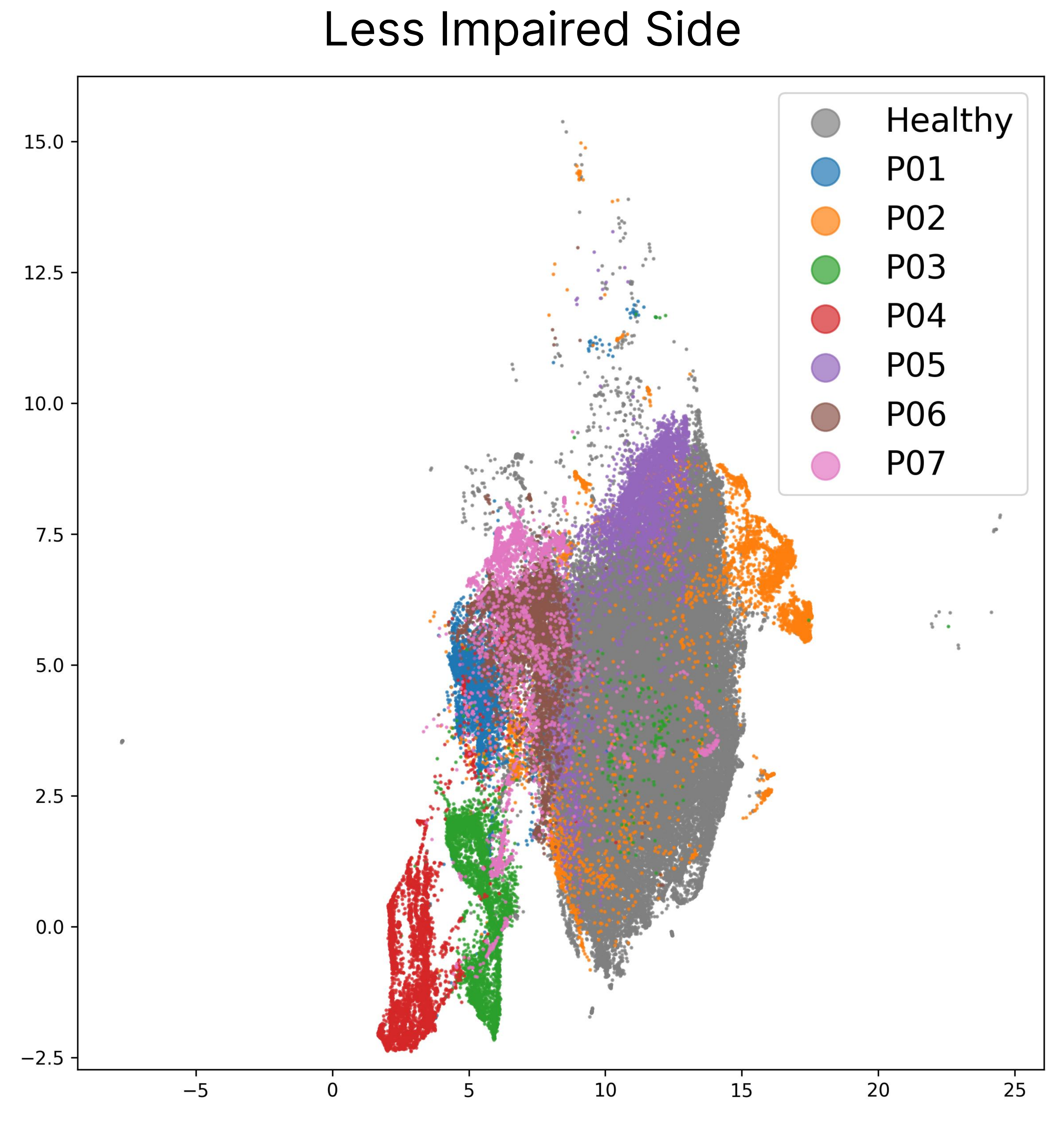}
    \caption{Healthy subject embeddings overlaid with joint-angle embeddings from each patient's less impaired side.}
    \label{fig:li_umap}
  \end{subfigure}
  \caption{UMAP embeddings of joint angles. Healthy embeddings are shown in all figures to provide a common reference for visual comparison. The separation between healthy and patient distributions is observed without using supervision or clinical labels.}
  \label{fig:umaps}
\end{figure*}

This comparison is included to motivate the choice of 3D pose estimation methods used in our joint angle and movement pattern analysis. Since reliable 3D ground-truth annotations are not available for the BBT dataset, we qualitatively evaluate temporal stability, robustness to hand and object occlusions, anatomical plausibility, and depth consistency.

\textbf{Evaluation setup:} We compare monocular methods for hand (WiLoR \cite{potamias2024wilor}), body (SMPLer-X \cite{cai2023smplerx}, PromptHMR \cite{wang2025prompthmr}), and body-hand (SAM 3D Body \cite{yang2025sam3dbody}) pose estimation. All methods were evaluated on identical video segments extracted from the same BBT recordings, using the same camera viewpoint, and pose estimates were generated without additional post-processing or temporal smoothing. Although the evaluated methods use different keypoint formats, they represent the same anatomical regions and were used only for a qualitative assessment of overall posture. Therefore, the comparison focuses on relative motion and stability rather than exact joint correspondence.

\textbf{Body pose comparison:} Fig. \ref{fig:body_pose} shows side-view visualizations of 3D keypoints during the grasp and transport phases of the Box and Block Test. All methods capture the overall motion of the arm associated with the task, and the front-view visualizations were similar for all the methods tested. However, qualitative differences are observed in depth consistency and joint articulation in the side-view visualizations. In particular, PromptHMR \cite{wang2025prompthmr} often produces an arm pose more elevated than suggested by the image, especially during the transport phase, indicating a tendency to over-extension in scenarios with depth ambiguity. In contrast, SMPLer-X \cite{cai2023smplerx} and SAM 3D Body \cite{yang2025sam3dbody} estimate the arm pose more closely aligned with the observed motion. These differences are important for joint angle analysis, where articulation bias can influence the estimated movement. Another difference observed was that SMPLer-X \cite{cai2023smplerx} shows reduced temporal consistency, with posture changes between the grasp and transport phases even when no corresponding change is visible in the images.

\begin{figure}[H]
  % \centering
  \includegraphics[width=\linewidth]{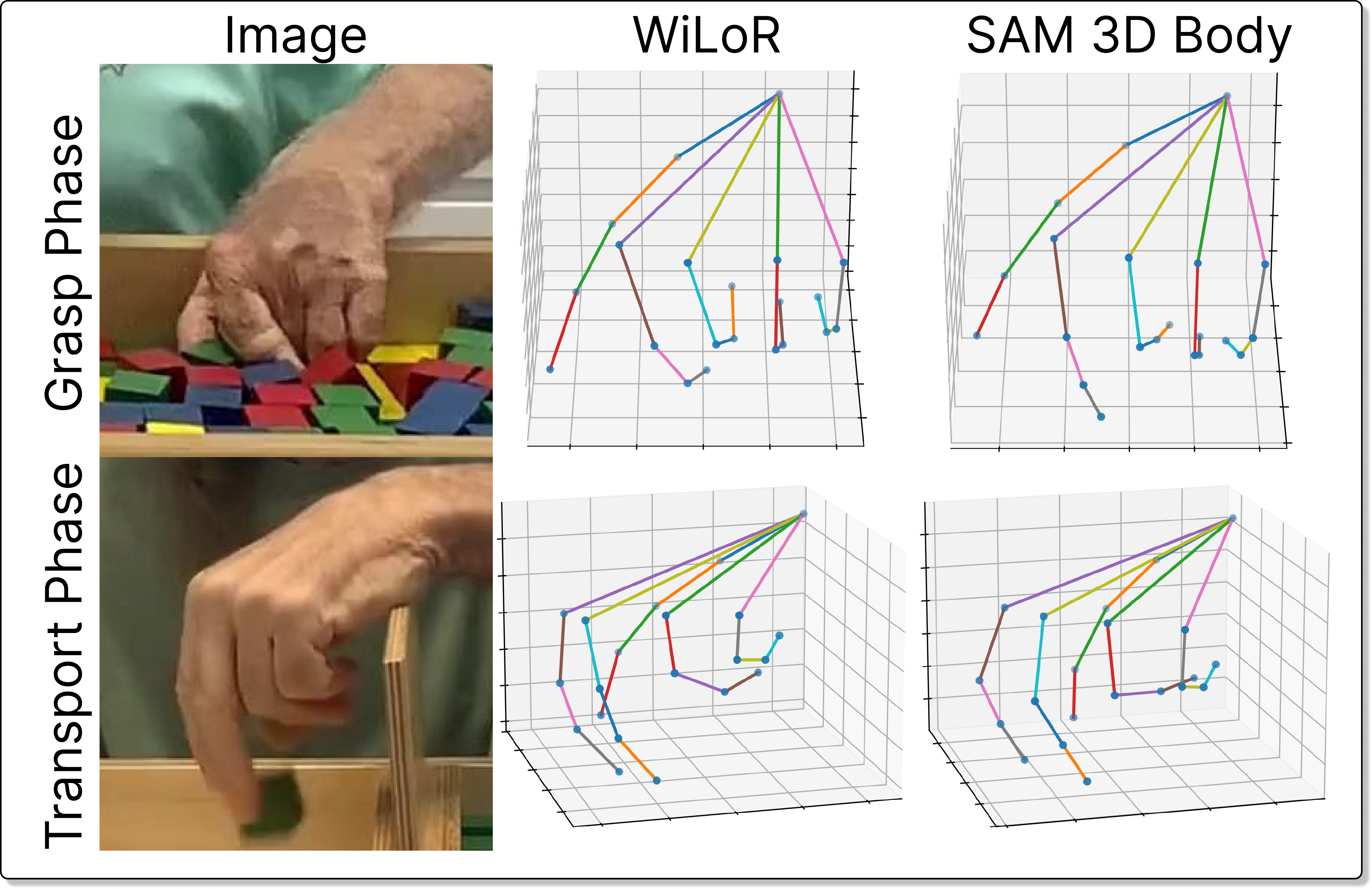}
  \caption{Comparison of 3D hand pose estimation methods during the grasp and transport phases of the Box and Block Test. Visualizations of 3D keypoints are shown for WiLoR and SAM 3D Body, illustrating differences in joint articulation relative to the observed hand pose in the image.}
  \label{fig:hand_pose}
  \vspace{-1em}
\end{figure}

\textbf{Hand pose comparison:} Fig. \ref{fig:hand_pose} shows visualizations of 3D keypoints during the grasp and transport phases of the Box and Block Test. WiLoR shows noticeable discrepancies in joint articulation. In several instances, the fingers that appeared strongly flexed in the image are reconstructed with a more extended pose. For example, in the transport phase of Fig. \ref{fig:hand_pose}, the ring finger is visually bent near a right angle but is reconstructed with a more extended joint configuration. In addition, both methods were trained on datasets consisting of healthy individuals and did not include complex poses, such as the grasp phase in Fig. \ref{fig:hand_pose}. In this image, the index finger overlaps on the middle finger, but is not represented in either model's prediction.

\textbf{Model selection:} Due to previously stated limitations, we chose SAM 3D Body \cite{yang2025sam3dbody} for both hand and body pose estimation. It produces body pose estimates that were more consistent with observed motion than those of SMPLer-X \cite{cai2023smplerx} and PromptHMR \cite{wang2025prompthmr}, and while it shares the same limitation in hand pose as WiLoR \cite{potamias2024wilor}, this limitation remains an open challenge for monocular hand pose estimation, to our knowledge. It is worth noting that SAM 3D Body also yielded the best or comparable performance on common benchmarks for body pose estimation (Tables \ref{tab:body_pose}).

\subsection{Differentiating Individuals Post Stroke and Healthy Individuals}

Having established a consistent method for estimating world-aligned 3D joint angles, we next investigate whether these angles capture meaningful differences between individuals post-stroke and healthy individuals. Specifically, we analyze upper-extremity joint angles extracted during the Box and Block Test and visualize their structure with a Uniform Manifold Approximation and Projection (UMAP) \cite{2018arXivUMAP}. We use UMAP due to its ability to preserve both local neighborhood structure and cluster relationships, which is important for interpreting distances between embeddings. This analysis is intended to assess whether stroke-related motor impairment causes deviations from normal movement patterns encoded by 3D joint angles.

\textbf{UMAP Configuration:} For each frame in each video (healthy subject and stroke patient), we extract 14 finger joint angles (three per finger and two for the thumb, excluding abduction/adduction of fingers and the carpometacarpal joint of the thumb). A PCA model \cite{santello1998postural} is fitted on finger joint angle samples of the healthy individuals to reduce the dimensionality of the fingers to 9 principal components (PCs), explaining over 90\% of the variance. These PC coefficients are used for patients as well. Additionally, we also extract angles of the shoulder (between upper arm and trunk vectors), elbow, wrist (between lower arm and palm vectors), and trunk (between trunk and gravity vectors). This results in a feature vector of 18 elements to represent each frame of the BBT. UMAP is fitted to the feature vectors using 20 neighbors and a minimum distance of 0.2 to reduce the dimensionality from 18 joint angles to two-dimensional coordinates for visualization.

\begin{table*}[t]
\centering
\begin{tabular}{ccccccccc}
\toprule
Patient & Gender & Age & FMA-UE & BBT Count (MI) & BBT Count (LI) & Stroke Subtype & Score (MI) & Score (LI) \\
\midrule
1 & M & 38 & 35 & 3.7 & 39.7 & Ischemic & 2.06 & 2.01 \\
2 & F & 70 & 58 & 13.7 & 30.0 & Hemorrhagic & 2.01 & 2.04 \\
3 & F & 41 & 43 & 4.3 & 22.7 & Hemorrhagic & 2.20 & 2.29 \\
4 & M & 68 & 30 & 3.3 & 11.7 & Hemorrhagic & 2.97 & 2.47 \\
5 & M & 52 & 52 & 17.0 & 41.0 & Ischemic & 2.19 & 1.39 \\
6 & M & 77 & 29 & 4.7 & 38.7 & Ischemic & 1.93 & 1.39 \\
7 & M & 58 & 48 & 15.0 & 30.3 & Ischemic & 1.52 & 1.53 \\
\bottomrule
\end{tabular}
\caption{Demographic and clinical characteristics of stroke patients. The table reports gender, age, FMA-UE scores where a lower score is more severe, average number of blocks moved across sessions for the more impaired (MI) and less impaired (LI) sides, stroke subtype, and unsupervised joint-angle deviation scores for each side. The score represents the KNN score as a normalized ratio to the mean KNN score among healthy frames, which is 16.15.}
\label{tab:patient_scores}
\vspace{-1.5em}
\end{table*}

\textbf{Joint-angle evaluation:} Since UMAP embeddings do not preserve absolute distances, all distance computations are performed in the original feature space. To quantify deviation from healthy movement patterns, we compared patient frames to healthy frames in the original PCA and arm joint-angle feature space. For each patient frame, we computed the mean Euclidean distance to its 15 nearest healthy frames and then averaged these distances across all frames for each side. The ratio between these distances and the mean pairwise distance among healthy frames was used to score patients' postures, as reported in Table \ref{tab:patient_scores}. 

\textbf{Qualitative comparison:} Patient scores were consistently higher than the healthy baseline (score > 1), with the more impaired side generally exhibiting larger deviations. This is visualized in the UMAP embeddings (Fig. \ref{fig:umaps}). Most importantly, the results yielded by our CV framework clearly provided additional information about the postures used by the participants to perform the task, which adds new dimensions to the BBT assessment. Note that we chose to use a frame-by-frame encoding of postural angles, which does not contain temporal information, because the raw BBT score is already a measurement of movement speed. There are several interesting observations. First, most patients had very different postures comparing the more impaired and less impaired side. While this is not surprising, we can observe that the postures of some less impaired sides are still different from those of healthy individuals. This indicates that many patients still use compensatory movement even for the less impaired limbs. Second, patients with similar BBT raw scores can be differentiated by posture patterns. For instance, Patients 2 and 7 (mildly impaired) have clinically indistinguishable BBT scores, but their joint angle embeddings show different clusters, potentially representing different compensatory strategies. The same argument can be made for Patients 1 (moderate) and 6 (severe). These qualitative comparisons demonstrate the potential of our framework to identify sub-groups of patients who may need different rehabilitation interventions despite scoring the same with the original BBT.

\section{Limitations}
There are several limitations that should be considered when interpreting the results. First, the number of individuals post-stroke is limited, with data collected from seven individuals across twenty sessions. This limits the generalizability of the observed movement patterns. Our ongoing research will expand the video dataset to involve a much larger sample size, and we will use both supervised and unsupervised machine learning algorithms to define sub-groups of patients. Second, we only performed frame-by-frame analysis of the postures, without any temporal information, which primarily quantifies compensatory movements. However, we are not excluding the possibility of alternative quantification based on our data. We are currently exploring other movement features, such as movement smoothness. Finally, our analysis is capturing all frames from videos. It is possible to integrate the present approach with our recent action phase classification algorithms \cite{Robinson2025MLSP}, enabling in-depth analysis of movement quality separately during grasping and transport phases of the BBT assessment.

\section{Conclusion}
In this paper, we introduce a computer vision-based framework for the analysis of upper-extremity movement during the Box and Block Test using world-aligned monocular 3D joint angles. By generating semantic segmentation maps of the test box to estimate camera pitch, we enforce gravity-aligned joint angle measurements that are robust to camera orientation and do not require depth sensors or additional calibration objects. This design allows movement quality to be analyzed from standard RGB video with minimal disruption to existing clinical assessment procedures. Using this framework, we analyzed joint-angle patterns from a dataset of 136 Box and Block Tests performed by healthy individuals and individuals post-stroke. Unsupervised embedding of finger, arm, and trunk joint angles revealed a separation between healthy movement patterns and stroke-related movement deviations, as well as distinct patterns within patients that share similar BBT scores. These results show that world-aligned 3D joint angles capture meaningful information about posture and compensatory behavior beyond standard time-based test scores. Such information can be used to assist therapists in identifying optimal post-stroke rehabilitation strategies during routine clinical visits. Furthermore, the proposed approach is not limited to individuals with stroke, but it can be extended to be used in other populations that frequently use BBT for UE assessment, such as individuals with Cerebral Palsy. Lastly, the proposed CV framework can also be extended to other time-based tests such as NHPT.

\section{Acknowledgements}
The authors thank the students Nina Castillo, Oscar Palacios Lara, David Marzan, and Bridget Waterman from the Physical Therapy Department of AdventHealth University for conducting all data collection, including the Box and Block Test, FMA-UE scores, and grip strength tests.

% \section{IRB Approval}

{\small
\bibliographystyle{ieee_fullname.bst}
\bibliography{egbib.bib}
}

\end{document}